\documentclass{llncs}
\usepackage[T1]{fontenc}
\usepackage{graphicx}
\usepackage{amsmath}
\usepackage{adjustbox}
\usepackage{threeparttable}
\usepackage{booktabs}
\usepackage{array}
\usepackage{tikz,hyperref}
\usepackage{xcolor}
\usepackage{geometry}
\usepackage{amssymb}
\usepackage{longtable}
\usepackage{subfigure}
\usepackage{color, soul}

\begin{document}
\title{Explainable Fuzzy GNNs for Leak Detection in Water Distribution Networks}

%

\newcommand{\orcidauthorA}{0009-0005-4748-4692}
\newcommand{\orcidauthorB}{0000-0002-4500-9098}
\newcommand{\orcidauthorC}{0000-0001-8935-9156}
\newcommand{\orcidauthorD}{0000-0001-9422-3157}
\newcommand{\orcidauthorF}{0000-0003-4661-7164}
\newcommand{\orcidauthorG}{0000-0001-8746-8826}

\author{Qusai Khaled\inst{1}\orcidauthorA{} \and
Pasquale De Marinis\inst{2}\orcidauthorC{} \and
Moez Louati\inst{3} \orcidauthorF{} \and
David Ferras\inst{4} \orcidauthorD{} \and
Laura Genga\inst{5} \orcidauthorG{} \and
Uzay Kaymak\inst{1}\orcidauthorB{}}
\authorrunning{Qusai Khaled et al.}
\institute{
Jheronimus Academy of Data Science, Eindhoven University of Technology, Eindhoven, The Netherlands
\email{qusai.khaled,U.Kaymak@ieee.org} 
\and
University of Bari Aldi Moro, Bari, Italy
\email{pasquale.demarinis@uniba.it}
\and
IHE Institute for Water Education, Delft, The Netherlands\\
\email{m.louati@un-ihe.org} 
\and
University of Cartagena, Cartagena, Spain
\email{david.ferras@upct.es} 
\and
School of Industrial Engineering, Eindhoven University of Technology, Eindhoven, The Netherlands
\email{l.genga@tue.nl}}
\maketitle
\begin{abstract}
Timely leak detection in water distribution networks is critical for conserving resources and maintaining operational efficiency. Although Graph Neural Networks (GNNs) excel at capturing spatial-temporal dependencies in sensor data, their black-box nature and the limited work on graph-based explainable models for water networks hinder practical adoption. We propose an explainable GNN framework that integrates mutual information to identify critical network regions and fuzzy logic to provide clear, rule-based explanations for node classification tasks. After benchmarking several GNN architectures, we selected the generalized graph convolution network (GENConv) for its superior performance and developed a fuzzy-enhanced variant that offers intuitive explanations for classified leak locations. Our fuzzy graph neural network (FGENConv) achieved Graph F1 scores of 0.889 for detection and 0.814 for localization—slightly below the crisp GENConv’s 0.938 and 0.858, respectively—yet it compensates by providing spatially localized, fuzzy rule-based explanations. By striking the right balance between precision and explainability, the proposed fuzzy network could enable hydraulic engineers to validate predicted leak locations, conserve human resources, and optimize maintenance strategies. The code is available at \href{https://github.com/pasqualedem/GNNLeakDetection}{https://github.com/pasqualedem/GNNLeakDetection}.
\end{abstract}

\section{Introduction}

Water distribution networks (WDNs) are vital infrastructures that support urban development and public health. Leakages in these systems lead to significant water loss, economic burdens, environmental impacts, and operational inefficiencies \cite{liu2024contrastive}. Globally, WDNs lose 32 billion cubic meters of treated water annually \cite{liemberger2006challenge}, while in developing countries losses could meet the annual water needs of 200 million people \cite{thornton2008water}. The financial costs of repairs and replacements amount to millions of dollars each year. Therefore, timely and accurate leak detection is crucial for efficient water resource management and minimizing operational disruptions.

Leak detection and localization methods can be broadly classified based on factors such as scale, accuracy, speed, cost, and efficiency, as well as their ability to detect small leaks. Hardware-based approaches, including ground-penetrating radar, tracer gas injection, and thermographic cameras, offer high accuracy and efficiency. But often come with significant costs and complexity. These methods excel in detecting leaks at specific locations but may struggle with large-scale network.\cite{kammoun2022leak}. Software-based methods, such as statistical, transient and machine learning methods rely on computational analysis from sensor data to infer leak locations. These methods generally offer high efficiency and fast detection times, with relatively lower costs compared to hardware-based methods. However, their performance depends on the availability and quality of sensor data, as well as the robustness and generalization ability of their underlying models. Unlike hardware-based methods, which provide direct physical detection, software approaches leverage inferred signals from transient responses or statistical patterns, making them highly scalable but sensitive to model assumptions, noise and network conditions \cite{bohorquez2022stochastic}.

    Recent studies have used a diverse range of machine learning techniques for leak detection and localization in WDNs. Traditional classifiers such as Support Vector Machines \cite{rashid2015wml} \cite{rajasekaran2024survey}\cite{guo2021leakage}, Bayesian classifiers \cite{soldevila2017leak}, genetic algorithms \cite{wu2010pressure} \cite{nasirian2013leakage}, Artificial Neural Networks (ANNs) \cite{mounce2006burst} \cite{fan2021machine}, and K-Nearest Neighbors \cite{ravichandran2021ensemble}\cite{tariq2022data} have all been applied to various aspects of the problem. In simpler pipeline systems, transient-based data analyzed through Convolutional Neural Networks (CNNs) have enabled precise leak localization \cite{bohorquez2020leak} \cite{bohorquez2022stochastic}, while steady-state sensor data have been used to pinpoint leaks at junctions in large-scale networks \cite{shekofteh2020methodology}. Additionally, CNNs have been employed to convert time-series pressure signals into images for enhanced performance \cite{javadiha2019leak} \cite{yu2024leak}. To address class imbalance challenges, unsupervised approaches like autoencoders have also been integrated with ANN frameworks \cite{fan2021machine}. More recent work has shifted towards supervised ensemble methods—including models like XGBoost, LightGBM, CatBoost, and deep neural networks—which, when trained on high-resolution sensor data, have achieved classification accuracies as high as 99.79\% \cite{lee2023machine}. Despite these advances, many approaches still rely on Euclidean data representations, which do not fully capture the intrinsic topological structure of WDNs.

These traditional architecture have shown to perform well in leak detection tasks with simple WDNs, but they tend to underperform in more complex, large-scale networks \cite{cui2024hierarchical}. To better capture the non-Euclideans nature of WDNs - characterized by irregrular graph like structures where nodes represent junctions and edges represent pipes - recent studies have turned to graph neural networks (GNNs). Common architectures such as graph convolutional networks (GCNs), graph attention networks (GATs), advanced GCN architectures, and hybrid models like SVM-CNN combined with graph-based localization have been explored. However, research remains limited when it comes to leveraging the graph structure of WDNs to develop explainable models for leak detection and localization.

Considering the significant role of explainability for the creation of trust-worthy machine learning models in water management \cite{richards2023rewards} and the need for trustworthiness for this innovations to make it to practice\cite{doorn2021artificial}\cite{hadjimichael2016machine}. Domain experts require not only high predictive accuracy but also explainability to validate model decisions before wasting human resources to the common pitfall of false positives. We propose an explainable GNN that synergies fuzzy logic with mutual information to provide semantic, rule-based interpretability while preserving strong predictive performance.  Our approach builds upon state-of-the-art GNN architectures and introduces a fuzzy-enhanced variant, which facilitates localized, post-hoc fuzzy rule-based explanations of sub-graph regions, all while maintaining semantic interpretability constraints and competitive performance.

The remainder of this paper is organized as follows. Section 2 reviews related work on leak detection in water distribution networks, with an emphasis on graph-based and machine learning approaches. Section 3 details our methodology, including the evaluation of various graph neural network architectures, data preprocessing and training protocols, and the development of an explainable fuzzy GNN framework. Section 4 presents the experimental results, comparing model performance in leak detection and localization tasks, and discussing the trade-offs between accuracy and interpretability. Finally, Section 5 concludes the paper and outlines directions for future research.

\section{Related work}
The use of graph structures in machine learning for leak detection has been found to accelerate the analysis of large complex networks \cite{shekofteh2020methodology}. GCNs emerged as the predominant GNN architecture for leak detection \cite{chen2023leak}\cite{brahmbhatt2023digital}\cite{csahin2023prediction}\cite{li2024convolutional}. Brahmbhatt et al. (2023) \cite{brahmbhatt2023digital} used GCNs to detect and localize leaks using pressure and flow data, creating digital twins for large WDNs. Their semi-supervised approach treated each pipe as a class to address limited monitoring stations. Similarly, Li et al. (2024) \cite{li2024convolutional} tackled limited monitoring by using pressure and flow data from two WDNs, achieving 90\% accuracy in high-density monitoring areas and 85\% in low-density areas. Their model integrated node and edge features through fusion strategies, highlighting the importance of optimized monitoring layouts. Şahin et al. (2023) \cite{csahin2023prediction} established a small-scale experimental setup to collect data for GCN training, outperforming SVM baselines in classifying nodes as leaky, non-leaky, or at-risk for early detection. Their work focused on detection rather than localization, demonstrating GNNs' superiority over traditional machine learning. Chen et al. (2023) \cite{chen2023leak} introduced an improved graph convolutional network (IGCN) using self-learning fully connected association graphs instead of fixed undirected graphs, eliminating the need for expert knowledge. Their model included bidirectional edges between nodes to represent potential causal relationships. Their leakage-detection algorithm incorporated data correlation of different labels through backpropagation, outperforming traditional autoencoder, CNN, and MLP-GCN models in two case studies.

Beyond GCNs, Wu et al. (2010) \cite{wu2010pressure} utilized graph attention network (GAT) for detecting leaks using simulated pressure and flow data, in their work they demonstrated GAT capabilities in detecting leaks in large scale WDNs of up to 858 nodes based on a grid system in Kentucky meant for supplying water for 6,400 customers. Researchers have also explored hybrid approaches combining graph-theory with conventional machine learning models. Shekofteh et al. (2020) \cite{shekofteh2020methodology} decomposed WDNs into sub-clusters then trained ANNs to detect and localize leaks using pressure sensor data from 5 simulated scenarios with demand and pressure uncertainty. Their methodology, tested on an actual network, identified leak locations as combinations of nodes and recommended acoustic methods for precise localization. Similarly, Kang et al. (2017) \cite{kang2017novel} proposed fusing one-dimensional CNN with SVM for leak detection alongside a graph-based localization algorithm. Their hybrid approach achieved 99.3\% accuracy with localization error below 3 meters in a real-world test bed under noisy conditions, outperforming both SVM and CNN when used individually.

Overall, integrating graph-based approaches has yielded significant benefits by leveraging the relational information embedded within the non-Euclidean structure of WDNs. These approaches have demonstrated consistent performance improvements over traditional machine learning methods, with accuracies reaching 94-98\% by explicitly modeling the topological relationships between nodes and pipes. However, a critical limitation persists across existing graph-based leak detection systems: while they excel at prediction tasks, they fail to harness the structural advantages of graph representations for interpretability purposes. Our work addresses this limitation by introducing a fuzzy GNN architecture aiming at maintaining competitive predictive performance while providing semantic, fuzzy rule-based explanations.

\section{Methodology}

To address leak detection and localization in WDNs, we leverage GNNs to capture the spatial and temporal dependencies in graph-structured data. We evaluate six GNN architectures—GCNConv~\cite{kipf2016semi}, SAGEConv~\cite{hamilton2017inductive}, GATConv~\cite{velivckovic2017graph}, GATv2Conv~\cite{brody2021attentive}, GENConv~\cite{li2020deepergcn}, and TransformerConv~\cite{shi2020masked}—each bearing different mechanisms for aggregating and updating node features, ranging from standard averaging to attention-based methods. These models are tested under identical conditions for two tasks: graph-level classification to detect leaks and node-level classification to pinpoint their locations.
Our experiments utilize the Hanoi Benchmark Network dataset (LeakDB)\cite{vrachimis2018leakdb}, a widely used open-source simulated dataset for testing leak detection models. It contains 1,000 scenarios, each representing a one-year time series with a 30-minute time step. All scenarios mimic the same network topology but differ in the number, location, and size of artificially induced leak. We construct a graph where nodes represent network components (e.g., junctions, reservoirs) and edges correspond to pipe connections. Node features represent the pressure values at the network's junctions while edge features represent the flow inside the pipes.

\subsection{GNN Architectures}
GNNs operate fundamentally on a graph \( G = (V, E) \), where \( V \) is the set of nodes and \( E \) the set of edges, with each node \( i \in V \) having a feature vector \( \mathbf{h}_i^{(l)} \) at layer \( l \). The core mechanism is message passing, where nodes aggregate information from their neighbors \( \mathcal{N}(i) \) and update their representations. We begin with GCNConv as the foundational architecture and progressively introduce more advanced variants, each building on this concept with distinct aggregation and update strategies.

\textbf{GCNConv}~\cite{kipf2016semi}: The Graph Convolutional Network layer serves as the baseline for GNNs by aggregating neighbor features using a normalized sum based on node degrees. In a graph, the degree of a node \( i \), denoted \( \deg(i) \), is the number of edges connected to it. The node feature computation at layer \( l+1 \) is given by:
    {\scriptsize
    \[
    \mathbf{h}_i^{(l+1)} = \sigma \left( \sum_{j \in \mathcal{N}(i) \cup \{i\}} \frac{1}{\sqrt{\deg(i) + 1} \sqrt{\deg(j) + 1}} \mathbf{W}^{(l)} \mathbf{h}_j^{(l)} \right),
    \tag{1}
    \]
    }
where \( \mathbf{W}^{(l)} \) is a learnable weight matrix, \( \sigma \) is an activation function (e.g., ReLU), and the normalization term in the denominator weights each neighbor’s contribution inversely proportional to the degrees of nodes \( i \) and \( j \) (with \( +1 \) accounting for self-loops). This process propagates and transforms node features across layers.

\textbf{SAGEConv}~\cite{hamilton2017inductive}: GraphSAGE enhances scalability by sampling neighbors and aggregating their features, typically using functions such as a
mean, LSTM or pooling,  before combining them with the central node's representation. The node feature computation at layer \( l+1 \) is given by:
    {\scriptsize
    \[
    \mathbf{h}_i^{(l+1)} = \sigma \left( \mathbf{W}^{(l)} \cdot \left[ \mathbf{h}_i^{(l)} \parallel \frac{1}{|\mathcal{N}(i)|} \sum_{j \in \mathcal{N}(i)} \mathbf{h}_j^{(l)} \right] \right),
    \tag{2}
    \]
    }
where \( \parallel \) denotes concatenation, allowing the model to preserve both the node's own information and that of its neighborhood.

\textbf{GATConv}~\cite{velivckovic2017graph}: The Graph Attention Network incorporates an attention mechanism to dynamically weigh neighbors' contributions based on feature relevance. The node feature computation at layer \( l+1 \) is given by:

{\scriptsize  
\[
\mathbf{h}_i^{(l+1)} = \sigma \left( \sum_{j \in \mathcal{N}(i)} \alpha_{ij}^{(l)} \mathbf{W}^{(l)} \mathbf{h}_j^{(l)} \right),
\tag{3}
\]
}
where attention coefficients \( \alpha_{ij}^{(l)} \) are computed as  
{\scriptsize \( \alpha_{ij}^{(l)} = \text{softmax}_j \big( \text{LeakyReLU} \big( \mathbf{a}^{(l)T} [\mathbf{W}^{(l)} \mathbf{h}_i^{(l)} \| \mathbf{W}^{(l)} \mathbf{h}_j^{(l)}] \big) \big) \) } where \( \mathbf{a}^{(l)} \) is a learnable attention vector. This allows GATConv to focus on significant neighbors adaptively.

\textbf{GATv2Conv}~\cite{brody2021attentive}:: An improvement on GATConv that enhances the attention mechanism by replacing static attention biases with dynamic, learnable functions. This allows attention weights to be computed based on both node features and their pairwise relationships, resulting in more expressive modeling of node interactions and improved performance while maintaining a similar update structure.

\textbf{GENConv}: designed to overcome the challenges of overfitting and vanishing gradients, the Generalized Graph Convolution (GENConv) aims to generalize message passing mechanisms and supports various aggregation functions like Softmax or PowerMean, as proposed in the DeepGCN paper \cite{wu2010pressure}.

{\scriptsize
\[
\mathbf{h}_i^{(l+1)} = \text{MLP} \left( \mathbf{h}_i^{(l)} + \text{AGG} \left( \{ \text{ReLU}(\mathbf{h}_j^{(l)} + \mathbf{e}_{ji}) + \epsilon : j \in \mathcal{N}(i) \} \right) \right)
\tag{4}
\]
}
where
 \( \text{MLP} \) is a Multi-Layer Perceptron applied to the final combined vector for transformation. \( \mathbf{e}_{ji} \) represents the features of the edge from node \( j \) to node \( i \).
\( \epsilon \) is a small constant often added for numerical stability.
\( \text{AGG} \) denotes the aggregation function operating over the set of processed messages from neighbors.

\textbf{TransformerConv2}~\cite{shi2020masked}:  
    Adapting the transformer architecture for graph data, this network employs self-attention over neighbors to update node representations. Notably, its aggregation follows the same structural form as GATConv in formula 3. The key difference is that while GATConv computes attention coefficients using a learnable attention vector and a LeakyReLU applied to concatenated transformed features, TransformerConv derives \( \alpha_{ij}^{(l)} \) via the transformer mechanism—employing queries, keys, and values—and can incorporate edge features. This yields a more dynamic attention computation that flexibly models pairwise relationships compared to the fixed parameterization in GATConv.

\subsection{Explainability Framework}
\label{sec:explainability_framework}

We propose an explainability framework that integrates GNN-based prediction with fuzzy logic and mutual information to provide semantic, rule-based interpretations of model decisions. The framework targets two main angles: first, identifying the critical subgraph region that most influences the GNN's prediction; and Secondly, deriving fuzzy rules from the identified subgraph to generate human-readable explanations of how specific features drive the model's decision.

\begin{figure}[t]
    \centering
    \includegraphics[width=1\linewidth]{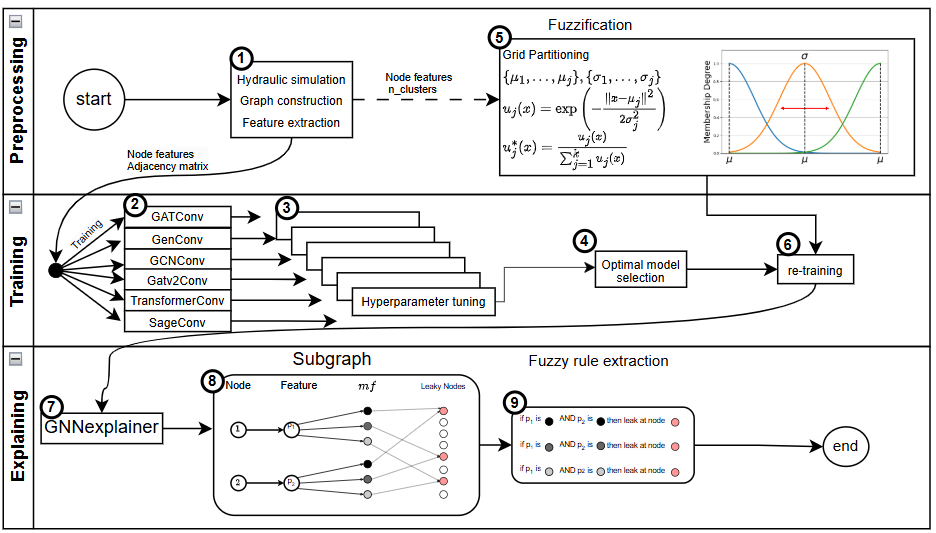}
    \caption{Explainability framework for GNN-based leak detection in water distribution networks, illustrating the sequential process: preprocessing, training, tuning, model selection, retraining with fuzzification, explanation via GNNExplainer, subgraph identification, and finally, fuzzy rule extraction for interpretable predictions.}
    \label{fig:exp}
\end{figure}

Figure~\ref{fig:exp} illustrates the proposed framework, which builds upon the model selection and training pipeline described earlier, and shows the explainability framework in alignment with the broader GNN modeling pipeline described earlier. Below, we detail the major steps of our approach.

From the preprocessing outputs in step 1, node features and the adjacency matrix are obtained. Then multiple GNN architectures are trained on the crisp dataset, and the best-performing model is identified based on leak detection and localization metrics as illustrated in steps 2, 3 and 4. Then the model with the optimal performance from step 4 is selected for re-training on fuzzified data to create the fuzzy variant FGENConv. A grid-partitioning fuzzification approach is implemented in step 5, dividing the data range of each numerical feature into three Gaussian membership functions, Low, Medium, and High. The means and standard deviations of these Gaussians are determined by evenly partitioning the feature space. Following the training of the fuzzy model in steps 7 and 8, the region of the graph contributing most to the model’s prediction is identified using GNNExplainer~\cite{ying2019gnnexplainer}; an explainability framework that maximizes the mutual information between the model’s prediction \( Y \) and a subgraph \( G_S \) of the original graph \( G \). In practice, it highlights a set of critical nodes (and their connecting edges) whose features most influence the final output. Finally, having localized the critical subgraph via GNNExplainer, we translate the influential nodes and their fuzzified features into human-readable rules. The last step aligns with an existing method from \cite{castellano2024integrating}, while also introducing two key distinctions:

\textbf{Node-Level Focus:} The extracted rules are tailored for node-level leak localization rather than graph-level classification, better addressing WDN leak localization needs where pinpointing faulty nodes is crucial.  
    
\textbf{Fuzzification Scheme:} We impose four semantic interpretability constraints to preserve the meaning associated with membership functions \cite{gacto2011interpretability}:  
    \textbf{Normality:} Each Gaussian set peaks at membership value 1.  
    \textbf{Convexity:} Unimodal Gaussians naturally satisfy convexity.  
    \textbf{Coverage:} The three sets collectively span the full feature range.  
    \textbf{Distinguishability:} Overlaps are constrained below a set threshold \cite{jin2000fuzzy} to maintain semantic interpretability.  

Once GNNExplainer identifies the contributing membership functions within the subgraph, the most influential node features are selected and combined via logical \texttt{AND} operators. This yields fuzzy rules of the example form:

\[
\small
\text{IF Pressure at Node 1 is high AND Pressure at Node 2 is low,}
\text{THEN Leak probability at Node } 5 \text{ is } 70\%.
\]

Such rules could provide a more intuitive for domain experts by enabling a semantic interpretation of the model’s prediction. The probability is obtained as the output of the sigmoid function, maintaining a probabilistic nature to reflect uncertainty in predictions.

\subsection{Experimental Configuration}
\begin{figure}[t]
    \centering
    \includegraphics[width=\linewidth]{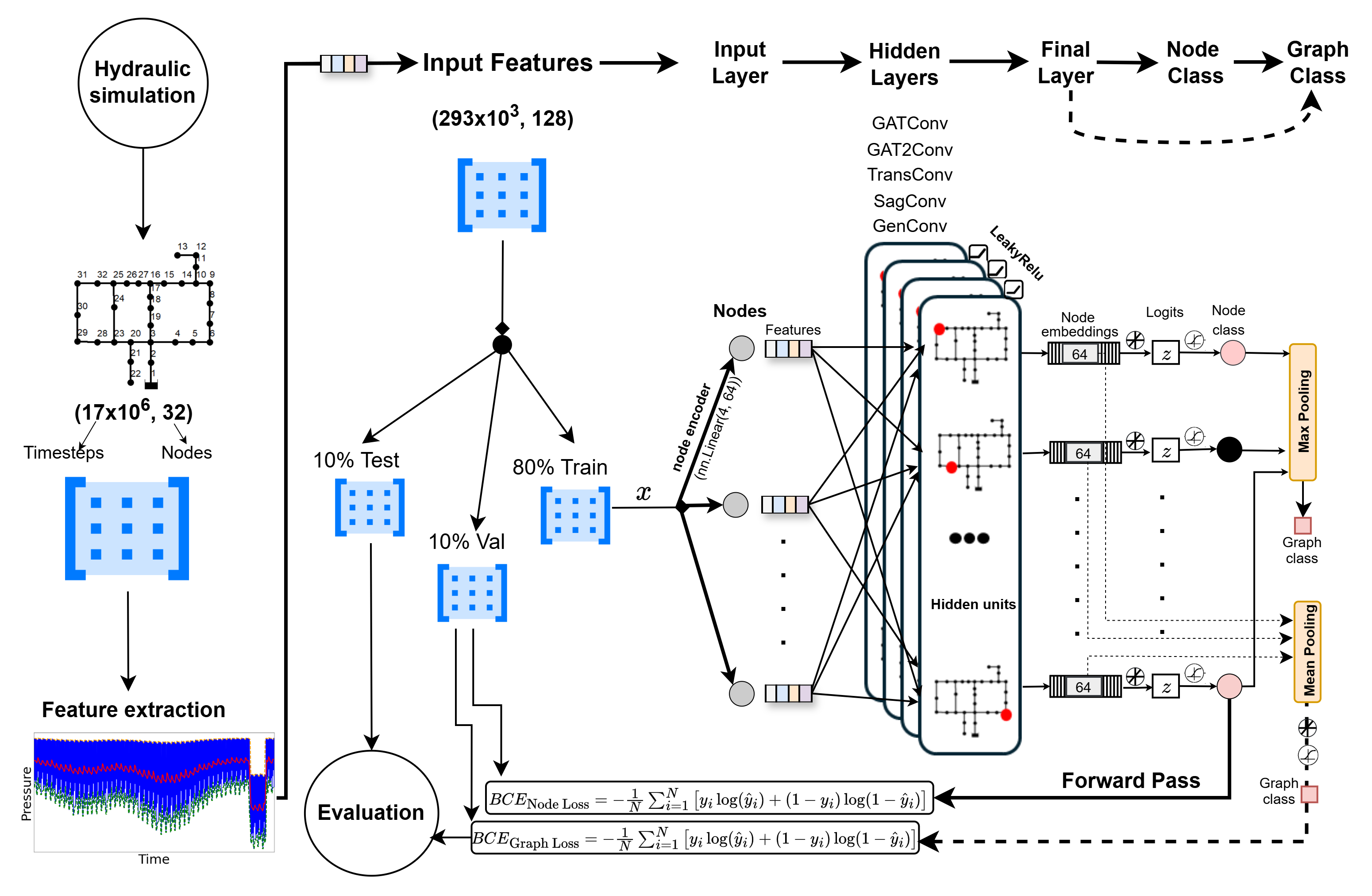}    \caption{Experimental configuration for leak detection and localization. Continuous arrows indicate the node-level leak localization path, while dashed arrows show the graph-level leak detection path.}
    \label{fig:model}
\end{figure}

Two task-specific models are trained from a shared GNN backbone, as shown in Figure \ref{fig:model}. The first is a node-level leak localization model, which processes node embeddings (solid arrows) through a linear layer with sigmoid activation. It is trained using node-level binary cross-entropy loss, and predictions are max-pooled to provide supplementary graph-level information. The second is a graph-level leak detection model, which aggregates node embeddings (dashed arrows) via mean pooling before applying a linear layer with sigmoid activation. This model is trained using graph-level binary cross-entropy loss.   Both models are optimized using the AdamW algorithm under identical experimental conditions.  

For feature extraction, raw pressure sensor data is preprocessed using a sliding window approach. Given a time-series signal \(X = [x_1, x_2, \ldots, x_T]\), windowed segments \(X_w^k = [x_k, x_{k+1}, \ldots, x_{k+W-1}]\) are generated using a window size \(W\) and stride \(S\). From each segment, statistical features such as the mean (\(\mu\)), standard deviation (\(\sigma\)), minimum, and maximum are computed to capture the network's operational state. The processed data is then used to construct a graph \(\mathcal{G} = (\mathcal{V}, \mathcal{E})\), where vertices \(\mathcal{V}\) represent the nodes of the WDN and edges \(\mathcal{E}\) represent the physical pipe connections between them. The adjacency matrix \(\mathbf{A}\) is generated based on the provided topology, with \(A_{ij} = 1\) indicating a direct connection between nodes \(i\) and \(j\). Each node \(i\) is assigned a feature vector \(\mathbf{h}_i\) derived from the preprocessed sensor data.

Model training is conducted with hyperparameters optimized via grid search, as detailed in Table~\ref{tab:hyperparams_horizontal}. We employ early stopping with a patience of 10 epochs and evaluate performance using \(F_1\)-scores for both graph- and node-level classification, ensuring a balanced assessment of precision and recall in leak detection tasks.

\begin{table}[b]
\centering
\scriptsize 
\caption{Hyperparameters and Experimental Configurations}
\label{tab:hyperparams_horizontal}
\renewcommand{\arraystretch}{1.1} 
\setlength{\tabcolsep}{6pt}      
\begin{tabular}{p{0.3\linewidth} p{0.3\linewidth} p{0.3\linewidth}}
\toprule
\textbf{Model Architecture} & \textbf{Training Protocol} & \textbf{Optimization} \\
\midrule
\textbf{GNN Layer Types:} \newline
GENConv, SAGEConv, GCNConv, TransConv, 
GATConv, GATv2Conv \newline
\textbf{Layer Depth:} 2, 4, 8 \newline
\textbf{Hidden Dimension:}  16, 32, 64 \newline
\textbf{Activation:} LeakyReLU \newline 
\textbf{Edge Handling:} Undirected \newline
&
\textbf{Learning Rate:} 0.001 \newline 
\textbf{Batch Size:} 512 \newline
\textbf{Epochs:} 500 \newline
\textbf{Patience:} 10 epochs \newline
\textbf{Normalization:}  LayerNorm \newline
\textbf{Task Mode:} \newline Graph or Node classification
&
\textbf{Optimizer:} AdamW \newline
\textbf{LR Scheduler:} \newline ReduceLROnPlateau  \newline
\textbf{Gradient Clipping:} \newline 1.0 (norm threshold) \newline
\textbf{Loss:} \newline BCEWithLogitsLoss  \\
\bottomrule
\end{tabular}
\end{table}

\section{Results and Discussion}
The results of the experiments, summarized in Table~\ref{tab:gnn_results}, compare the previously mentioned GNN architectures—including GATConv, GCNConv, SAGEConv, TransformerConv, and GENConv— as well as the fuzzy variant FGENConv. These models were evaluated across two tasks: (1) leak detection (a graph-level classification problem) and (2) leak detection combined with leak localization (a joint graph and node classification task). The evaluation metrics include graph F1 score, node F1 score, test loss, and runtime.

\subsection{Leak Detection/Graph Classification}
For the leak detection task, all models were trained under identical experimental conditions with a consistent range of network depths and hidden layer dimensions.

\textbf{GENConv} emerged as the top performer, achieving a mean Graph F1 score of \textbf{0.938} with a correspondingly low test loss of \textbf{0.116}, while maintaining a moderate runtime (ranging from 1730 to 6747 seconds). This implies GENConv is able to capture the spatial dependencies in WDNs.

\textbf{FGENConv}, the fuzzy variant, recorded a Graph F1 score of \textbf{0.889} with a higher test loss (0.154) and a runtime of 4205 seconds. Implying slightly lower performance and higher average run time. The anticipated performance trade-off, estimated around 5\% is argued to be insignificant in light of the gained explainability advantage.

Other architectures, such as GATConv, GATv2Conv, and TransformerConv, provided comparable performance with Graph F1 scores in the range of 0.927--0.934. Their runtime, however, varied more widely (from 1271 up to 7046 seconds), indicating that runtime efficiency may be sensitive to the specific architecture and hyperparameter configurations (e.g., the wider range of layers used by GATConv).

\subsection{Leak Localization/Node Classification}
In the more challenging joint task of leak detection and localization, both graph-level and node-level predictions were evaluated.

\textbf{GENConv} again achieved the best performance with a Graph F1 score of \textbf{0.855} and a Node F1 score of \textbf{0.805}, along with the lowest test loss of \textbf{0.028}. These results highlight its robustness in simultaneously capturing global network features and localized node-level anomalies.

\textbf{FGENConv} attained a Graph F1 of \textbf{0.814} and a Node F1 of \textbf{0.758}. While scores are modestly lower than those of GENConv, they remain within the 5\% trade-off estimated from the graph classification model.

\textbf{GATv2Conv} (Graph F1: 0.835, Node F1: 0.779), \textbf{TransformerConv} (0.808, 0.746), and \textbf{SAGEConv} (0.783, 0.717) showed reasonable performance in the joint task of leak detection and localization, though their scores trailed behind the top-performing

The runtime for the models in the joint task varied, with GENConv and FGENConv maintaining competitive runtimes (1440--2776 seconds for GENConv and a fixed 6683 seconds for FGENConv), underscoring also the balance between computational efficiency and interpretability in addition to the accuracy interpretability trade off. The significant increase in run time is likely due to the expanded feature space resulting from fuzzification.

Overall, GENConv delivered the highest predictive accuracy for both leak detection and localization, while its fuzzified version, FGENConv, showed a slight drop in performance. Nonetheless, it provided explainable outputs via fuzzy rule-based interpretations. This trade-off aligns with our design goals, where the added semantic interpretability supports water utility operators in making more informed maintenance and resource allocation decisions.

Furthermore, the comparative analysis reveals that the choice of GNN architecture is critical: while some architectures GATConv and TransformerConv perform robustly, they may incur longer runtimes. The stark underperformance of GCNConv in the joint task highlights that not all architectures readily extend to combined graph and node classification problems in the context of WDNs.

It is important to note that the F1-scores for graph-level tasks did not exceed 0.94, likely due to class imbalance, which realistically mirrors the infrequent occurrence of leaks in WDNs. The drop in F1-scores for node-level classification is even more pronounced, as the imbalance is exacerbated at the finer granularity—only a small subset of nodes are affected by leaks across the entire network. This significant imbalance at the node level accounts for the lower performance in node classification compared to the graph-level tasks.

\begin{table}[t]
\centering
\scriptsize
\begin{threeparttable}
  \caption{Aggregated performance of GNN models on leak detection (graph) and leak detection + localization (node) tasks. F1 Scores and losses refer to the best run score out of 20 runs with different hyperparameter configuration per model.}
  \label{tab:gnn_results}
  
  \begin{tabular}{l@{\hspace{1em}}c@{\hspace{1em}}c@{\hspace{1em}}c@{\hspace{1em}}c@{\hspace{1em}}c}
    \toprule
    \multicolumn{6}{l}{\textbf{Leak Detection Models} (Graph classification)} \\
    \midrule
    \textbf{Model} & \textbf{Layers} & \textbf{Hidden Size} & \textbf{Graph F1} & \textbf{Test Loss} & \textbf{Runtime (s)} \\
    \midrule
    GATConv	        & 2--8  & 16--64 & 0.935 & 0.118 & 1271--7046  \\
    GATv2Conv	      & 2--8  & 16--64 & 0.934 & 0.119 & 2156--9289  \\
    GCNConv	        & 2--8  & 16--64 & 0.934 & 0.118 & 1279--3700  \\
    GENConv	        & 2--8  & 16--64 & \textbf{0.938} & 0.116 & 1730--6747  \\
    SAGEConv	      & 2--8  & 16--64 & 0.936 & 0.117 & 1106--4619  \\
    TransformerConv	& 2--8  & 16--64 & 0.934 & 0.119 & 1544--10410 \\
    \midrule
    FuzzyGENConv        & 4  & 64 & 0.889 & 0.154 & 4205--4205 \\
    \bottomrule
  \end{tabular}

  \vspace{1em} 
  
\begin{tabular}{l@{\hspace{1em}}c@{\hspace{1em}}c@{\hspace{1em}}c@{\hspace{1em}}c@{\hspace{1em}}c@{\hspace{1em}}c}
  \toprule
  \multicolumn{7}{l}{\textbf{Lelak Detection and Localization Models} (Graph and Node Classification)} \\
  \midrule
  \textbf{Model} & \textbf{Layers} & \textbf{Hidden Size} & \textbf{Graph F1} & \textbf{Node F1} & \textbf{Test Loss} & \textbf{Runtime (s)} \\
  \midrule
  GATConv	        & 2--8  & 16--64 & 0.845	& 0.786 & 0.031 & 1506--16143 \\
  GATv2Conv	      & 2--8  & 16--64 & 0.856	& 0.810 & 0.029 & 1237--12098 \\
  GCNConv	        & 2--8  & 16--64 & 0.113	& 0.106 & 0.062 &  745--5635 \\
  GENConv	        & 2--8  & 16--64 & \textbf{0.858}	& \textbf{0.811} & 0.028 & 1083--8115 \\
  SAGEConv	      & 2--8  & 16--64 & 0.862	& 0.774 & 0.033 &  807--4216 \\
  TransformerConv	& 2--8  & 16--64 & 0.868	& 0.783 & 0.031 & 1266--9214 \\
  \midrule
  FuzzyGENConv        & 4 & 64 & 0.814 & 0.758 & 0.032 & 6683--6683 \\
  \bottomrule
\end{tabular}
  
  \begin{tablenotes}
    \footnotesize
    \item \textbf{Note:} Bold indicates top performance within each task.
  \end{tablenotes}
\end{threeparttable}
\end{table}

\section{Conclusion}

This paper presented an explainable fuzzy graph neural network framework for detecting and localizing leaks in water distribution networks (WDNs). By integrating fuzzy logic and mutual information, our approach was able to capture spatial-temporal dependencies while maintaining interpretability. We identified GENConv as the top-performing GNN architecture for combined leak detection and localization tasks. We also introduced FGENConv, a fuzzy-enhanced variant offering rule-based explanations for improved interpretability, aiding experts and reducing false positives. This marks the first integration of fuzzy logic and GNNs in leak detection, preserving both semantic interpretability and predictive performance. Though results are promising, validation in larger, diverse WDNs is needed. Future work could incorporate edge features, pipe-level localization, and expand to larger scale networks.

\begin{credits}
\subsubsection{\ackname} \small This publication is part of the project Innovation Lab for Utilities on Sustainable Technology and Renewable Energy project (ILUSTRE), which is partly financed by the Dutch Research Council (NWO).

\subsubsection{\discintname}
The authors have no competing interests to declare that are
relevant to the content of this article..
\end{credits}
%
%
%

\bibliographystyle{splncs04}
\bibliography{ref}

\end{document}